\documentclass[10pt,twocolumn,letterpaper]{article}
\usepackage[pagenumbers]{wacv}
\usepackage{comment}

\usepackage{amsmath,amssymb,mathtools,booktabs,multirow,graphicx,xcolor,makecell,stfloats,placeins,float}
\definecolor{wacvblue}{rgb}{0.21,0.49,0.74}
\usepackage[breaklinks,colorlinks,allcolors=wacvblue]{hyperref}
\title{When Evidence Vanishes: Distilling Visual-Evidence Dependency in Medical Vision-Language Models}
\author{\parbox{0.96\textwidth}{\centering\fontsize{12}{14}\selectfont
Qizhen Lan\textsuperscript{1,*}, Yiling Ma\textsuperscript{2,*}, Ruiqi Chen\textsuperscript{3}, Mengchen Fan\textsuperscript{4},\\
Xi Xiao\textsuperscript{4}, Lin Zhao\textsuperscript{5}, Liang He\textsuperscript{6}, Baocheng Geng\textsuperscript{4},\\
Pulin Che\textsuperscript{4}, Tianyang Wang\textsuperscript{4}, Xiaoqian Jiang\textsuperscript{1,$\dagger$}\\[5pt]
{\fontsize{10}{12}\selectfont
\textsuperscript{1}The University of Texas Health Science Center at Houston\quad
\textsuperscript{2}Yale University\\
\textsuperscript{3}University of Michigan\quad
\textsuperscript{4}University of Alabama at Birmingham\\
\textsuperscript{5}Northeastern University\quad
\textsuperscript{6}ETH Zurich\\[4pt]
\href{mailto:Qizhen.Lan@uth.tmc.edu}{Qizhen.Lan@uth.tmc.edu}}}}
\date{}
\begin{document}
\maketitle
\begingroup
\renewcommand{\thefootnote}{\fnsymbol{footnote}}
\footnotetext[1]{Qizhen Lan and Yiling Ma contributed equally.}
\footnotetext[2]{Corresponding author: Xiaoqian Jiang.}
\endgroup
\begin{abstract}
Compact medical vision-language models (VLMs) can answer clinical questions
correctly without reliably using the image, relying instead on question wording or learned priors. Standard knowledge distillation copies a teacher's answer preference on the original image, but it leaves unchecked whether that preference is conditional on the visual evidence required by the question. We study this missing target as \emph{medical visual evidence dependency}: support for the correct answer should decrease when the matched evidence is absent. We introduce \emph{Evidence-Conditioned Dependency Distillation} (ECDD), which pairs each training image with a control matched to the evidence level and distills the teacher's drop in answer margin across the pair. A lower bound objective transfers positive dependency by penalizing only shortfalls, while original-image supervision preserves task accuracy. The same formulation applies when the evidence is a queried clinical region, a surgical instrument, or whole-image visual content. Across three medical VQA settings, ECDD keeps original-image accuracy competitive with distillation and alignment baselines while making
answer support more sensitive to evidence removal. Under the DME headline objective, for example, the positive-case margin response to target rather than random masking rises from $0.786$ with KD to $2.023$ with ECDD, while original-image accuracy differs by $0.11$ points. These results suggest that medical VLM distillation should transfer both the teacher's answer preference and its dependence on visual evidence.
\end{abstract}

\section{Introduction}
\label{sec:intro}
\begin{figure*}[t]
\centering
\includegraphics[width=0.95\textwidth]{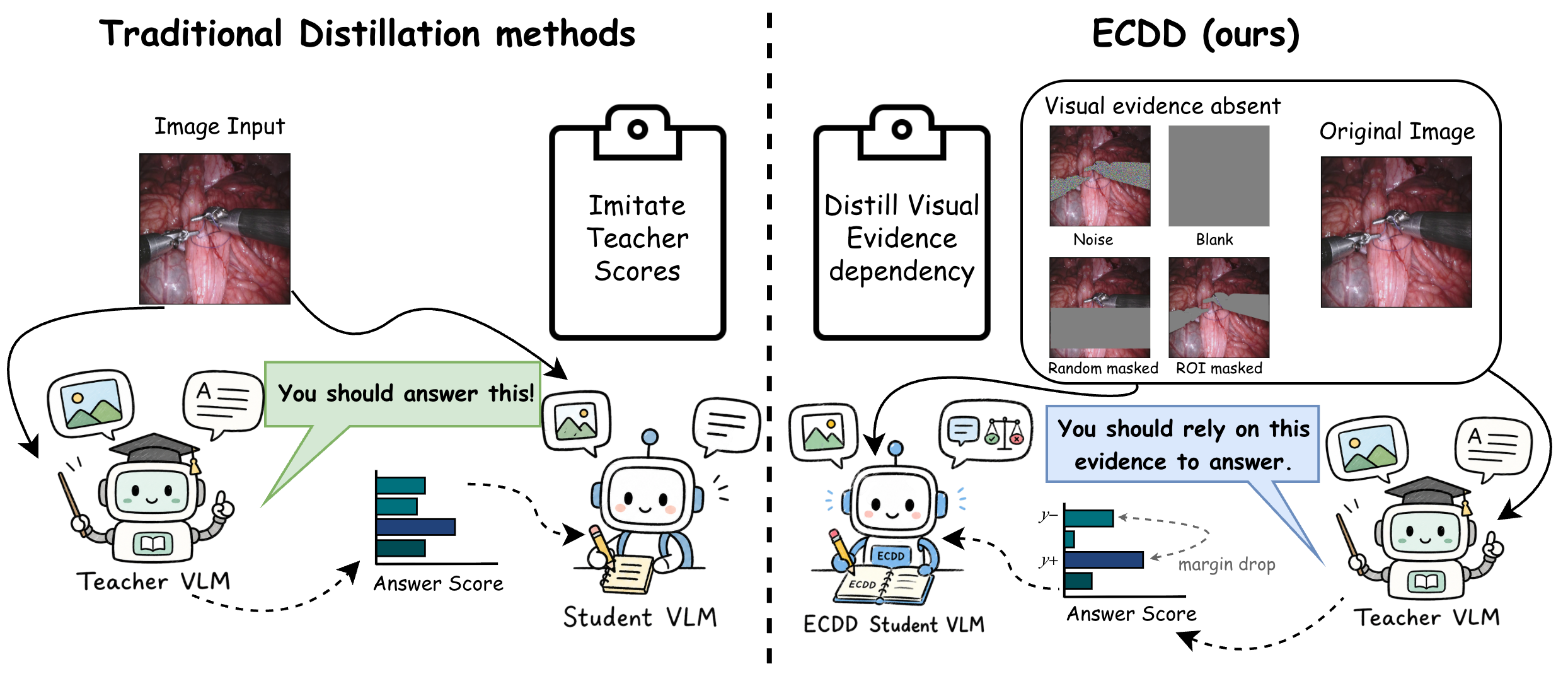}
\caption{\textbf{Static answer distillation vs. evidence-dependency distillation.}
Standard distillation copies the teacher's answer scores on the original image only, leaving the evidence-removed endpoint unseen. ECDD instead compares the original image $x$ with an evidence-ablated control $A(x)$ and distills the teacher's dependency gap, measured as the drop in the margin between the gold answer and the hardest negative. Here, $y^+$ is the gold answer and $y^-$ is the strongest wrong answer.}
\label{fig:intro}
\end{figure*}

Compact medical vision-language models (VLMs) are useful when medical visual question answering (VQA) must run under local compute, latency, privacy, or hardware constraints. Knowledge distillation (KD) is a natural path for this setting. A large teacher can provide supervision for a smaller student, following classical knowledge distillation~\cite{hinton2015distilling} and recent multimodal distillation methods such as LLaVA KD~\cite{cai2025llava}. Efficient multimodal models, mobile VLMs, and small VLMs make this direction practical in vision-language systems~\cite{jin2024efficient,chu2023mobilevlm,zhou2024tinyllava}. Medical VQA therefore raises the research question that motivates this paper. After a compact student learns a teacher's answer preference, has it also learned whether that preference depends on the medical visual evidence required by the question? Original image accuracy alone cannot answer this question. As illustrated in Fig.~\ref{fig:intro}, standard distillation transfers a static teacher score at the original image, whereas ECDD distills how answer support changes when the relevant visual evidence is removed.

This question matters because VQA accuracy can be supported by language priors rather than image evidence~\cite{goyal2017making,agrawal2018don,vo2026vision}. In medical VQA, these priors can be clinically plausible. Question wording, anatomy, modality, disease prevalence, and repeated templates may already favor one answer. A model can therefore select the correct option while giving nearly the same support after the relevant visual evidence is removed. We call the desired property \emph{medical visual evidence dependency}: answer support should decrease when the visual evidence required by the question is removed. We measure this change with a matched evidence-absent control that keeps the question, prompt, and candidate answers fixed. The evidence unit is not always the whole image. Some questions depend on a lesion region. Others depend on task-specific visual evidence such as surgical instruments.
Existing distillation methods do not directly train this dependency. Answer distillation transfers logits or probabilities at the observed image. Alignment methods transfer features, attention, or medical visual alignment signals~\cite{chang2026enhancing,chang2025focus}. Response structure methods such as Switch KD compare structured visual responses~\cite{sun2026switchkd}. Medical VQA methods such as KDAR use adaptive teacher soft labels to reduce language prior effects~\cite{peng2025overcoming}. These methods improve what the student predicts under the observed image, but they do not supervise how answer support changes when the evidence is absent. This leaves two gaps. \textbf{\emph{G1. Pointwise supervision misses evidence dependence.}} A loss defined only at the original image cannot identify the student's response at an evidence absent endpoint. \textbf{\emph{G2. Medical evidence is level specific.}} The relevant evidence may be a whole image, a queried lesion region, or a task-specific visual object. The dependency target should therefore be matched to the evidence level required by the question.

We propose \textbf{\emph{Evidence-Conditioned Dependency Distillation} (ECDD)} to train this missing target. ECDD pairs each original image with a control matched to the evidence level and measures the teacher's margin drop from original to control. The student is trained with a lower bound loss that penalizes only dependency shortfall, while original image supervision is retained to preserve task performance. This changes the object of distillation from a static answer preference at one image to an evidence-conditioned response across two controlled endpoints. The same formulation applies to global image controls, question-specified region masks, and task-specific visual evidence masks.

\paragraph{Contributions.}
We make three contributions.
\begin{itemize}
  \itemsep0.2em
  \item We formulate \textbf{\emph{medical visual evidence dependency}} as a missing target in compact medical VLM distillation. This separates original image answer correctness from whether answer support remains conditional on the evidence required by the question.
  \item We introduce \textbf{\emph{ECDD}}, a dependency distillation objective that transfers the drop in the teacher's margin from the original image to the evidence absent control with a lower bound constraint.
  \item We provide an evaluation across evidence levels, spanning localized clinical evidence, task-specific visual objects, and global controls without visual evidence. ECDD improves evidence-dependency diagnostics while maintaining competitive original-image accuracy in the evaluated compact student models.
\end{itemize}

\section{Related Work}
\label{sec:related}
\paragraph{Medical VQA benchmarks and evidence diagnostics.}
Medical VQA has become a common testbed for multimodal clinical reasoning. Benchmarks span radiology, pathology, and broader biomedical images, including VQA-RAD~\citep{lau2018dataset}, ImageCLEF-VQA-Med~\citep{ben2019vqa}, PathVQA~\citep{he2020pathvqa}, PMC-VQA~\citep{zhang2023pmc}, and OmniMedVQA~\citep{hu2024omnimedvqa}. These benchmarks measure answer accuracy, but closed-set accuracy alone does not show whether answer support depends on the visual evidence required by the question. The relevant evidence can also appear at different levels~\citep{feng2026trimrg}. It may be a whole image, a retinal region~\citep{feng2026tfvsf}, a surgical instrument, or a histopathology finding. Our evaluation therefore complements global medical VQA benchmarks with evidence-removal controls matched to the queried evidence level: region-of-interest (ROI) masking for DME region VQA, instrument-pixel masking for RoboSurg-VQA, and global visual-evidence-absent controls for PathVQA.

\paragraph{Language priors and counterfactual VQA evaluation.}
A long line of VQA work has shown that models can exploit answer-frequency
biases, question templates, and language priors, motivating question-only
baselines, changing-prior splits, adversarial evaluation, shuffled-image
controls, and blank-image diagnostics~\citep{goyal2017making,agrawal2018don,ramakrishnan2018overcoming,huang2026bayesvqa,zhu2026language}.
RUBi learns a question-only branch to reduce unimodal answer bias~\citep{cadene2019rubi};
CSS synthesizes counterfactual questions and images by masking critical words or
objects~\citep{chen2020counterfactual}; and CF-VQA uses a causal decomposition to
separate language effects from multimodal evidence~\citep{niu2021counterfactual}.
These protocols test whether prediction accuracy survives when the image is
irrelevant, degraded, shuffled, or absent. In medical VQA, such diagnostics are especially useful because repeated clinical question templates can make some answers
partly predictable from text alone. ECDD builds on this counterfactual diagnostic perspective. Instead of treating the evidence-absent input only as
a post hoc probe or reweighting examples by unimodal bias, ECDD transfers the
teacher's one-sided margin change across a question-matched medical-evidence
contrast. The distilled object is therefore a cross-endpoint dependency rather
than a static answer distribution or a generic debiasing weight.

\paragraph{Distillation for compact and medical VLMs.}
Knowledge distillation transfers a large teacher's behavior to a smaller
student~\citep{hinton2015distilling,lan2026performance,fan2026dcd,ding2026dprd,lan2025clockdistill,lan2025acamkd} and has been widely adopted for efficient
multimodal models. Recent multimodal large language model (MLLM) distillation methods transfer answer distributions, hidden states, visual token relations, or multimodal alignment signals. For example, LLaVA KD distills multimodal distributions and relation structure~\citep{cai2025llava}, LLaVA MoD uses MoE distillation for efficient small MLLMs~\citep{shu2025llava}. LLAVADI studies practical design choices for MLLM distillation~\citep{xu2024llavadi}. In medical VLMs, MedAlign distills visual alignment and attention knowledge~\citep{chang2026enhancing}, while KDAR uses teacher soft labels and adaptive reweighting to reduce VQA language-prior effects~\citep{peng2025overcoming}. Closely related to our setting, Switch KD goes beyond standard logit matching by using visual switch distillation and a Dynamic Bidirectional Logits Difference (DBiLD) loss to align informative teacher and student probability regions~\citep{sun2026switchkd}. These methods improve compact or medically aligned VLMs, but they primarily
regularize predictions, representations, or response structure under a given
input condition. ECDD instead distills a cross-condition evidence-removal
functional, defined by how the margin between the gold answer and the hardest negative changes
between the original image and a matched evidence-absent control.

\section{Method}
\label{sec:method}

\begin{figure*}[t]
  \centering
\includegraphics[width=0.88\textwidth]{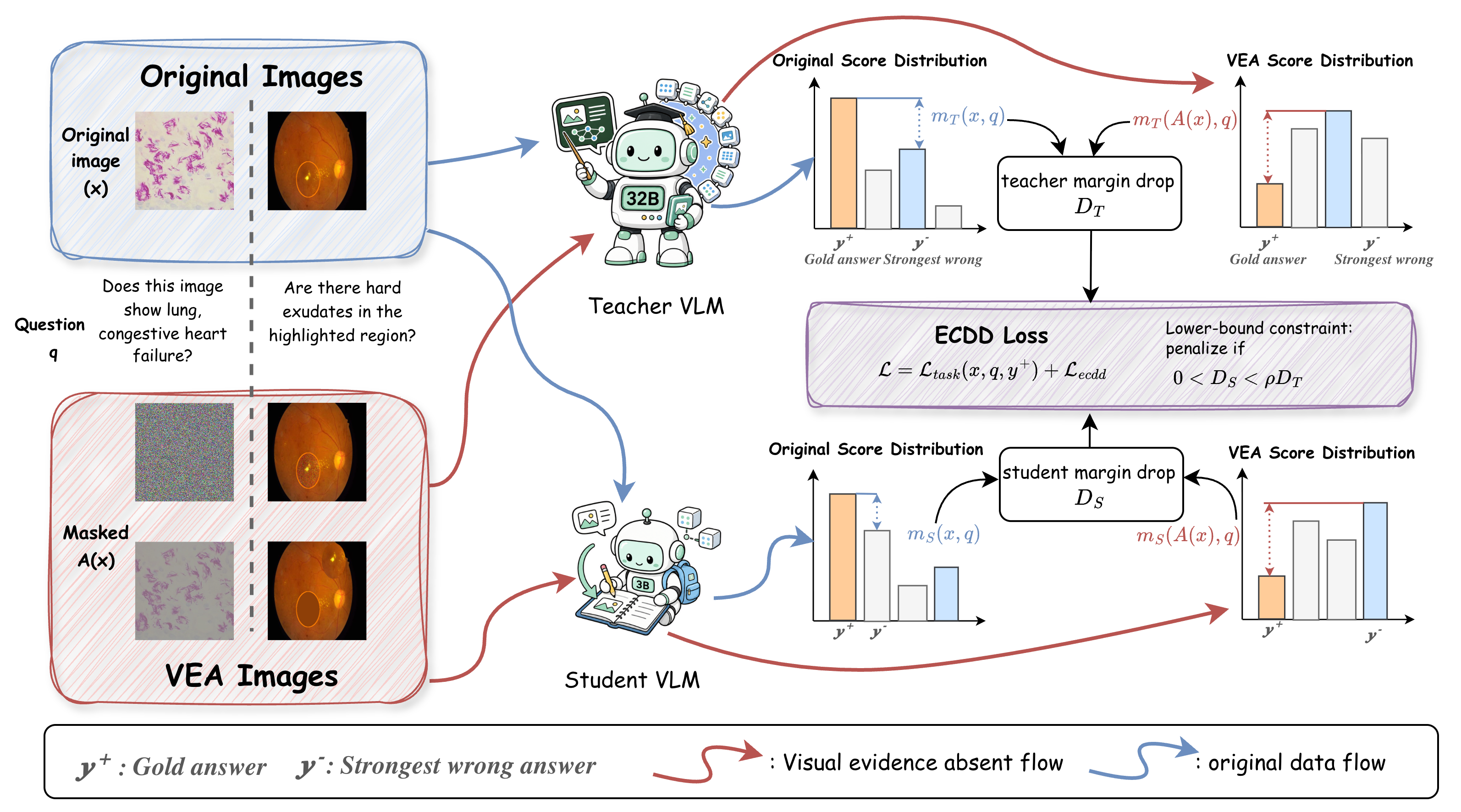}
\caption{\textbf{Overview of ECDD.}
ECDD pairs each original image with a level-matched control in which the relevant visual evidence is absent. The teacher's original-to-control margin drop is cached as a dependency target, and the student is trained with original-image supervision plus a one-sided dependency constraint that penalizes only shortfalls in evidence dependence.}

  \label{fig:ecdd_workflow}
\end{figure*}

ECDD trains the relation that the introduction identifies as missing from pointwise distillation. As shown in \Cref{fig:ecdd_workflow}, the target goes beyond the teacher's answer preference at the original image. It is the change in teacher answer support when the visual evidence required by the question is removed. ECDD first constructs a contrast matched to the evidence level, measures the drop in the margin between the gold answer and the strongest wrong answer, and trains the student to preserve positive teacher dependency while retaining original-image behavior.

\subsection{Problem setting and candidate scoring}
\label{sec:problem}

We consider closed-set medical VQA. Each example contains a medical image $x$, a question $q$, and a fixed candidate set $\mathcal{C}=\{a_1,\ldots,a_K\}$. The gold answer is $y^+\in\mathcal{C}$. The student $S$ is a compact VLM, and the teacher $T$ is a larger VLM used during training.

All methods use the same option-letter interface. The prompt asks for a single option letter. For model $M\in\{T,S\}$ under image condition $z$, let $r_M(k\mid z,q)$ denote the vocabulary logit assigned to the option token for candidate $a_k$. We read this logit at the first answer position. 
We normalize over the fixed-candidate set:

\begin{equation}
  p_M(k\mid z,q)
  =
  \frac{\exp r_M(k\mid z,q)}
  {\sum_{j=1}^{K}\exp r_M(j\mid z,q)} .
  \label{eq:candidate_prob}
\end{equation}

We use the log candidate probability $s_M(k\mid z,q)=\log p_M(k\mid z,q)$ as the candidate score below. The prediction is $\operatorname*{arg\,max}_{k} s_M(k\mid z,q)$. This interface keeps training and evaluation comparisons about the learning objective rather than free-form answer parsing.

The same interface covers binary and multiple-choice VQA. When $|\mathcal{C}|=2$, the strongest wrong answer is the other option. When $|\mathcal{C}|>2$, it is selected from all wrong candidates. Because teacher and student score scales need not be calibrated, ECDD uses margin drops through a scaled lower bound objective rather than exact raw margin matching.

\subsection{Evidence contrast matched to the queried level}
\label{sec:evidence_contrast}

Evidence dependency cannot be defined from the original image alone. Let $A$ be a visual-evidence-ablation operator for an example. The operator is matched to the evidence level required by $q$. It keeps the question, candidate answers, prompt template, resolution, and preprocessing fixed. It changes only the visual evidence whose presence is needed to answer the question. 

For global visual evidence, $A(x)=\tilde{x}$ is a visual-evidence-absent (VEA) image control. For lesion-region evidence, $A(x)$ masks the question-specified retinal region. For task-specific object evidence, $A(x)$ masks the visual object required by the question, such as surgical instrument pixels. The ordered paired evidence contrast is

\begin{equation}
  \mathcal{P}_A(x)=\bigl(x, A(x)\bigr).
  \label{eq:paired_contrast}
\end{equation}

This paired endpoint measures whether answer support changes when the relevant
evidence is absent while all language inputs remain fixed. Because medical
evidence may be global, regional, or scene-local, the control endpoint is
matched to the queried evidence unit (\textbf{G2}). We interpret the resulting
contrast as evidence dependence under the specified intervention~\cite{zafar2026beyond}.

\subsection{Evidence dependency margin}
\label{sec:margin}

Top-1 answer changes are too coarse for dependency supervision. A model can keep the same predicted answer while moving its decision boundary. At each endpoint $z$, we therefore define the margin between the gold answer and the strongest wrong answer as

\begin{equation}
  m_M(z,q)
  =
  s_M(y^+\mid z,q)
  -
  \max_{k\in\mathcal{C}\setminus\{y^+\}} s_M(k\mid z,q).
  \label{eq:margin}
\end{equation}
The maximizing wrong candidate may differ between $x$ and $A(x)$, but the candidate set is fixed. This margin records boundary movement even when Top-1 accuracy does not change.

The evidence dependency functional is the original to control margin drop:
\begin{equation}
  D_M(x,q;A)
  =
  m_M(x,q)-m_M(A(x),q).
  \label{eq:functional}
\end{equation}

A positive $D_M$ means that removing the matched visual evidence lowers support for the gold answer relative to the strongest wrong answer. A value near zero means that the model gives similar support at the two endpoints.

For localized evidence experiments, we also report a target-specificity diagnostic:
\begin{equation}
  D_M(x,q;A_{\mathrm{target}})
  -
  D_M(x,q;A_{\mathrm{rand}}).
  \label{eq:specificity}
\end{equation}
Here $A_{\mathrm{target}}$ removes the queried evidence, and $A_{\mathrm{rand}}$ removes a same-area random or distractor region. This diagnostic separates target-evidence removal from a generic occlusion response. It is not part of the default ECDD training loss.

\subsection{Why pointwise distillation misses dependency}
\label{sec:theory}

Pointwise distillation observes only the original image endpoint, whereas evidence dependency is a two-endpoint property. Let $\Phi_M(x,q)$ be any teacher or student observable measured only at the original image, such as logits, probabilities, features, attention maps, or a scalar margin. A pointwise distillation loss has the form
\begin{equation}
  \mathcal{L}_{\mathrm{pt}}
  = \ell\!\left(\Phi_S(x,q),\Phi_T(x,q)\right).
  \label{eq:pointwise}
\end{equation}
This loss may improve original-image prediction, but it gives no training signal for the student margin at $A(x)$.

This is a direct identifiability limitation of pointwise supervision. Two
students can agree on $\Phi(x,q)$, and therefore receive the same
$\mathcal{L}_{\mathrm{pt}}$, while assigning different margins to $A(x)$. Their
dependency functionals then differ through the unconstrained term
$m_S(A(x),q)$. Pointwise matching can improve prediction at the observed image,
but it cannot specify the student's response to evidence removal. This is why
ECDD uses paired supervision.

\subsection{Lower-bound ECDD loss}
\label{sec:objective}

Exact dependency matching is unnecessary because the teacher margin response can vary in scale across examples and model families. ECDD instead treats positive teacher dependency as a lower bound target. Let
\begin{equation}
  D_T^+(x,q;A)=\max\{D_T(x,q;A),0\},
  \label{eq:positive_dependency}
\end{equation}
and clip extreme positive teacher drops before scaling:
\begin{equation}
  \widetilde{D}_T^+(x,q;A)
  =
  \min\{D_T^+(x,q;A),\tau\},
  \label{eq:clipped_dependency}
\end{equation}
where $\tau$ is a fixed dependency cap. The ECDD loss is
\begin{equation}
  \mathcal{L}_{\mathrm{ECDD}} =
  \operatorname{ReLU}\!\left(
    \rho \widetilde{D}_T^+(x,q;A) - D_S(x,q;A)
  \right)^2 ,
  \label{eq:ecdd_loss}
\end{equation}
where $\operatorname{ReLU}(u)=\max\{u,0\}$ and $\rho\in[0,1]$ scales the amount of teacher dependency requested from the student. The loss penalizes only dependency shortfall. It does not force the student to copy every teacher fluctuation, and it does not punish a student whose margin drop already exceeds the scaled target.

For localized presence tasks, the ECDD term is applied only when the
queried evidence is labeled present. For DME, we further retain only teacher
targets for which the teacher is correct on the original image and $D_T^+$
exceeds a fixed reliability threshold. Other examples retain original-image task
supervision without being forced to lose support under masking. This filter
prevents teacher errors or negligible endpoint fluctuations from becoming
dependency targets; it uses only training labels and cached teacher scores.

\subsection{Overall objective and training instantiation}
\label{sec:overall_objective}

For single-stage training, ECDD combines original-image task learning, ordinary
answer-distribution distillation, and evidence-dependency transfer:
\begin{equation}
  \mathcal{L}
  =
  \mathcal{L}_{\mathrm{CE}}
  + \lambda_{\mathrm{KD}}\mathcal{L}_{\mathrm{KD}}
  + \lambda_{\mathrm{ECDD}}\mathcal{L}_{\mathrm{ECDD}} .
  \label{eq:loss}
\end{equation}
Here $\mathcal{L}_{\mathrm{CE}}=-\log p_S(y^+\mid x,q)$ and
$\mathcal{L}_{\mathrm{KD}}=\mathrm{KL}(p_T(\cdot\mid x,q)\|p_S(\cdot\mid
x,q))$ are evaluated only on the original image. Only
$\mathcal{L}_{\mathrm{ECDD}}$ compares $x$ with $A(x)$.

Dataset-specific training schedules and loss weights are specified in
\Cref{app:training_shared}. The reported PathVQA model uses ECDD as a
post-SFT regularizer with an original-image reference term, rather than the
single-stage schedule used for the Qwen-3B DME headline comparison.

Teacher dependency values and original-image teacher distributions are cached
before ECDD training. During training, the student is scored under both $x$ and
$A(x)$ to form $D_S(x,q;A)$. At inference, ECDD uses only the original image and
the same candidate-letter scoring interface as the supervised student.

\section{Experiments and Results}
\label{sec:experiments}

The experiments test whether ECDD preserves original-image task performance while increasing medical visual evidence dependency under controls matched to the evidence level. We evaluate lesion-region evidence in DME VQA~\citep{tascon2022consistency}, surgical instrument evidence in RoboSurg-VQA~\citep{zhang2026robosurg}, and global visual-evidence-absent (VEA) behavior in PathVQA~\citep{he2020pathvqa}. Original-image Top-1 accuracy is the task metric. Evidence-removal behavior is reported as a diagnostic of answer-support dependency.

\subsection{Experimental Setup}
\label{sec:setup}
\paragraph{Models.}
Unless stated otherwise, experiments use Qwen2.5-VL-Instruct~\citep{bai2025qwen25vl} with a 3B student (Qwen-3B) and a 32B teacher (Qwen-32B). On DME, we also test InternVL3.5-Instruct~\citep{wang2025internvl3} with a 2B student (InternVL-2B) and a 38B teacher (InternVL-38B). This second model family examines the same protocol under another compact student and teacher pair. All experiments use the fixed-candidate scoring interface defined in \Cref{eq:candidate_prob}.

\paragraph{Baselines.}
We compare ECDD with supervised fine-tuning and five distillation or alignment
baselines using matched backbones and a common candidate-scoring protocol.
Supervised fine-tuning (SFT) learns from gold answers on original images, whereas
KD matches the teacher's original-image candidate distribution~\citep{hinton2015distilling}.
Within this shared protocol, MedAlign~\citep{chang2026enhancing},
A3Tune~\citep{chang2025focus}, DBiLD from Switch KD~\citep{sun2026switchkd}, and
Beta-KD~\citep{sun2026uncertainty} use their respective original-image
objectives; ECDD adds the paired dependency target.

\paragraph{Datasets and evidence controls.}
We evaluate three evidence levels: lesion evidence in DME VQA~\citep{tascon2022consistency}, surgical instrument evidence in RoboSurg-VQA~\citep{zhang2026robosurg}, and global visual-evidence-absent behavior in PathVQA~\citep{he2020pathvqa}. DME and RoboSurg-VQA use localized controls with same-area random masks; PathVQA uses a global VEA control plus gray, noise, blur, block-shuffle, and image-swap diagnostics. A control is always scored against the original label: it probes whether answer support survives evidence removal and is not assigned a new clinical ground truth. The localized controls test question-matched region or object evidence, whereas the PathVQA control tests dependence on the image as a whole. \Cref{app:dataset_details,app:training_details} give filtering rules, split statistics, and implementation details.

\paragraph{Training.}
All adapted models use LoRA while the vision encoder remains frozen.
SFT uses original-image cross-entropy, while KD-style baselines add their
original-image distillation or alignment losses. Dataset- and backbone-specific
ECDD settings are listed in \Cref{app:training_shared}. For the Qwen-3B
DME headline comparison, we jointly optimize task supervision,
answer-distribution distillation, and ECDD in one stage with
$\lambda_{\mathrm{ECDD}}=0.02$, $\rho=0.5$, and $\tau=8$.
The DME ablations use the same task objective and optimization rule.
PathVQA instead uses one SFT epoch followed by one ECDD continuation epoch
with $\lambda_{\mathrm{KD}}=0$, $\lambda_{\mathrm{ECDD}}=0.05$,
$\rho=1$, $\tau=8$, and an original-image reference KL of weight $0.1$.
Its baseline comparisons therefore describe the complete training
instantiation, including the additional stage and reference regularization.
The ECDD term is active when the
relevant evidence is labeled present and the teacher provides a reliable
original-to-control response. Target selection and sensitivity analyses are
reported in \Cref{app:training_dme_regimes,app:sensitivity}.

\paragraph{Metrics.}
\begin{itemize}
  \setlength{\itemsep}{0pt}
  \setlength{\parsep}{0pt}
  \setlength{\topsep}{2pt}
  \item \textbf{Task performance.} \emph{Orig. Acc.} is Top-1 accuracy on the
  original image.
  \item \textbf{Localized dependency.} \emph{Evid. Gap} and \emph{Rand. Gap}
  are original-accuracy decreases under target-evidence removal and matched
  random/distractor controls; Rand. Gap should remain near zero. \emph{Specific
  $\Delta m$} subtracts the latter margin drop from the former. We average
  positive cases for presence ($\Delta m_+$) and all cases for localization.
  \item \textbf{Global dependency.} PathVQA \emph{VEA Gap} and
  $\Delta m_{\mathrm{VEA}}$ are the analogous original-to-VEA accuracy and
  gold-margin changes.
\end{itemize}
We interpret dependency diagnostics together with original-image accuracy.
Full formulas and secondary metrics are provided in \Cref{app:metric_details}.

\begin{table*}[!t]
  \centering
  \footnotesize
  \setlength{\tabcolsep}{3.6pt}
  \renewcommand{\arraystretch}{1.05}
  \resizebox{\textwidth}{!}{%
  \begin{tabular}{@{}llcccccccc@{}}
    \toprule
    \multirow{2}{*}{\textbf{Task}} & \multirow{2}{*}{\textbf{Method}}
      & \multicolumn{4}{c}{\textbf{Qwen-3B}}
      & \multicolumn{4}{c}{\textbf{InternVL-2B}} \\
    \cmidrule(lr){3-6} \cmidrule(lr){7-10}
    & &
    Orig Acc $\uparrow$ &
    Evid. Gap $\uparrow$ &
    Rand. Gap $\to 0$ &
    Specific $\Delta m \uparrow$ &
    Orig Acc $\uparrow$ &
    Evid. Gap $\uparrow$ &
    Rand. Gap $\to 0$ &
    Specific $\Delta m \uparrow$ \\
    \midrule

    \multirow{9}{*}{\textbf{\makecell[l]{Lesion\\presence}}}
      & Teacher
      & 69.39 & 3.81 & 2.18 & 3.094
      & 81.26 & 16.99 & -1.85 & 4.396 \\
      & Base
      & 46.19 & 1.85 & 3.16 & 0.300
      & 77.67 & 17.32 & 2.29 & 2.750 \\
      & SFT
      & 62.96 & 3.05 & -0.87 & 0.990
      & 71.35 & 7.52 & -4.90 & 1.330 \\
      & KD~\cite{hinton2015distilling}
      & \textbf{80.50} & 3.81 & \textbf{-0.11} & 0.786
      & 77.78 & 12.85 & -5.45 & 1.410 \\
      & DBiLD~\cite{sun2026switchkd}
      & 80.28 & 4.68 & \textbf{-0.11} & 1.050
      & 78.87 & 14.60 & -5.12 & 1.830 \\
      & MedAlign~\cite{chang2026enhancing}
      & 74.62 & -0.65 & \underline{0.65} & 0.683
      & \underline{84.86} & 23.75 & -1.09 & \underline{2.571} \\
      & A3Tune~\cite{chang2025focus}
      & 79.19 & 2.61 & -0.76 & 0.799
      & \textbf{85.62} & \underline{24.29} & \textbf{0.00} & 2.288 \\
      & Beta-KD~\cite{sun2026uncertainty}
      & 78.87 & \underline{10.57} & -2.40 & \underline{1.362}
      & 79.96 & 15.03 & -4.36 & 2.015 \\
      & \textbf{ECDD (ours)}
      & \underline{80.39} & \textbf{16.88} & 0.87 & \textbf{2.023}
      & 83.77 & \textbf{24.95} & \underline{-0.87} & \textbf{3.950} \\

    \midrule

    \multirow{8}{*}{\textbf{\makecell[l]{Lesion\\localization}}}
      & Base
      & 56.33 & 6.67 & 9.67 & -0.080
      & 66.67 & 17.00 & 2.00 & 1.240 \\
      & SFT
      & 64.33 & -3.00 & -4.67 & \underline{0.370}
      & 69.00 & -7.33 & 2.00 & -0.650 \\
      & KD~\cite{hinton2015distilling}
      & 68.00 & 2.00 & -4.00 & 0.190
      & 78.00 & 6.33 & 1.33 & 0.420 \\
      & DBiLD~\cite{sun2026switchkd}
      & 69.33 & 5.00 & \textbf{0.00} & 0.220
      & \underline{78.67} & 5.33 & \underline{0.67} & 0.490 \\
      & MedAlign~\cite{chang2026enhancing}
      & 59.33 & \textbf{12.33} & \underline{-0.67} & 0.343
      & 72.33 & \textbf{13.67} & \textbf{-0.33} & \textbf{0.951} \\
      & A3Tune~\cite{chang2025focus}
      & \textbf{70.33} & 5.33 & 2.33 & 0.175
      & 77.00 & 8.00 & \textbf{0.33} & 0.474 \\
      & Beta-KD~\cite{sun2026uncertainty}
      & 65.33 & 0.00 & -5.67 & \textbf{0.428}
      & 77.33 & -5.00 & 1.33 & -0.480 \\
      & \textbf{ECDD (ours)}
      & \underline{70.00} & \underline{8.67} & \textbf{0.00} & 0.340
      & \textbf{80.00} & \underline{12.67} & 1.67 & \underline{0.580} \\

    \bottomrule
  \end{tabular}%
  }
  \caption{\textbf{DME localized evidence results.}
  Lesion presence has $n{=}918$ test questions; lesion localization has $n{=}300$. Specific $\Delta m$ is the target-minus-matched-random/distractor margin response, averaged over positive cases ($\Delta m_+$) for presence and all cases for localization. Rand. Gap is ranked by distance to zero. The Qwen-3B presence KD row uses the task-matched recipe in \Cref{tab:dme_target_ablation}. Bold and underline mark the best and second-best trained rows; Teacher and Base are unranked diagnostic references.}
  \label{tab:dme_roi}
\end{table*}

\subsection{Main Results}
\label{sec:main_results}

Across the three evidence levels, ECDD keeps original-image accuracy competitive while increasing the response to removing the evidence required by the question. Across seeds 7/17/27, the evaluated ECDD recipes improve mean dependency
response over original-image KD on all three datasets with comparable or
higher mean original accuracy (\Cref{tab:multiseed}). The PathVQA rows use
the staged recipe described in \Cref{app:training_shared}; their gains do not
isolate the dependency loss from training schedule and reference regularization.

\paragraph{DME localized evidence.}
\label{sec:dme_roi}
On Qwen-3B lesion presence, the positive-case target-minus-random response rises
from $0.786$ with matched KD to $2.023$ with ECDD. Original-image accuracy
differs by $0.11$ points ($80.50\%$ for KD and $80.39\%$ for ECDD). Compared with
DBiLD, ECDD differs by
$+0.11$ pp in original accuracy (95\% CI $[-0.87,1.09]$) and improves the ROI
gap and ROI--random response by $12.20$ and $0.97$.

On InternVL-2B presence, ECDD gives the largest evidence gap ($24.95$) and
target-specific response ($3.950$), with original-image accuracy $1.85$ points
below A3Tune. The gap falls from $17.32$ in the untuned Base to $7.52$ after
SFT, then rises to $24.95$ with ECDD, a pattern consistent with weakened and
partially recovered pretrained evidence sensitivity. On Qwen-3B lesion
localization, ECDD has a zero distractor gap and original-image accuracy within
$0.33$ points of A3Tune. On InternVL-2B, ECDD has the highest accuracy
($80.00\%$); MedAlign produces a larger response at $7.67$ points lower
accuracy.

\begin{table*}[!t]
  \centering
  \small
  \setlength{\tabcolsep}{4.8pt}
  \renewcommand{\arraystretch}{1.05}
  \begin{tabular}{@{}lccccc@{}}
    \toprule
    \textbf{Method}
    & Orig Acc $\uparrow$
    & Rand. Acc $\uparrow$
    & Evid. Gap $\uparrow$
    & Evid. $\Delta m_+ \uparrow$
    & Evid.-Rand. $\Delta m_+ \uparrow$ \\
    \midrule

    Base
    & 68.00
    & 68.67
    & -2.00
    & 0.328
    & 0.290 \\

    \midrule

    SFT
    & 61.00
    & 45.67
    & 27.33
    & 1.210
    & 0.511 \\

    KD~\cite{hinton2015distilling}
    & \underline{72.67}
    & \underline{51.33}
    & \underline{36.00}
    & 1.495
    & 0.589 \\

    Beta-KD~\cite{sun2026uncertainty}
    & 65.33
    & 46.00
    & 32.00
    & 1.698
    & \underline{0.704} \\

    DBiLD~\cite{sun2026switchkd}
    & 68.67
    & 47.00
    & 35.00
    & 1.660
    & 0.676 \\

    MedAlign~\cite{chang2026enhancing}
    & 72.00
    & \textbf{53.33}
    & 33.00
    & \underline{1.788}
    & 0.685 \\

    A3Tune~\cite{chang2025focus}
    & \textbf{73.00}
    & 50.67
    & 34.67
    & 1.189
    & 0.547 \\
    \midrule

    \textbf{ECDD (ours)}
    & \textbf{73.00}
    & 50.00
    & \textbf{39.33}
    & \textbf{1.811}
    & \textbf{0.738} \\

    \bottomrule
  \end{tabular}
  \caption{\textbf{Surgical instrument evidence on RoboSurg-VQA Q8}
  ($n{=}300$). Evidence denotes EndoVis2017 instrument pixels. Rand. Acc uses a same-area random mask. Evid. Gap, Evid. $\Delta m_+$, and Evid.-Rand. $\Delta m_+$ measure instrument-specific evidence dependence. Bold and underline mark the best and second-best trained rows; Base is an unadapted reference and is not ranked.}
  \label{tab:robosurg_q8}
\end{table*}

\begin{table*}[!t]
  \centering
  \small
  \setlength{\tabcolsep}{4.6pt}
  \renewcommand{\arraystretch}{1.05}
  \begin{tabular}{@{}lcccccc@{}}
    \toprule
    \textbf{Method}
      & Orig Acc $\uparrow$
      & VEA Acc $\downarrow$
      & VEA Gap $\uparrow$
      & $\Delta m_{\mathrm{VEA}} \uparrow$
      & Ret. $m_{\mathrm{VEA}} \downarrow$
      & ECE / Brier $\downarrow$ \\
    \midrule

    Base
      & 62.17
      & 56.72
      & 5.44
      & 0.120
      & 0.694
      & 8.78 / 47.06 \\

    \midrule

    SFT
      & \underline{88.55}
      & 82.27
      & \underline{6.28}
      & \underline{0.751}
      & 0.727
      & 2.14 / \underline{17.07} \\

    KD~\cite{hinton2015distilling}
      & 86.79
      & 81.17
      & 5.62
      & 0.408
      & 0.694
      & 8.84 / 21.34 \\

    Beta-KD~\cite{sun2026uncertainty}
      & 86.91
      & 81.95
      & 4.97
      & 0.604
      & 0.699
      & \underline{2.10} / 19.37 \\

    MedAlign~\cite{chang2026enhancing}
      & 84.62
      & 79.30
      & 5.32
      & 0.439
      & 0.663
      & 7.28 / 23.65 \\

    DBiLD~\cite{sun2026switchkd}
      & 84.00
      & \underline{78.67}
      & 5.32
      & 0.450
      & \underline{0.661}
      & 6.53 / 24.28 \\

    A3Tune~\cite{chang2025focus}
      & 85.22
      & 80.37
      & 4.85
      & 0.457
      & 0.692
      & 5.32 / 22.25 \\

    \midrule

    \textbf{ECDD (ours)}
      & \textbf{89.35}
      & \textbf{69.10}
      & \textbf{20.26}
      & \textbf{3.863}
      & \textbf{0.158}
      & \textbf{1.38} / \textbf{15.19} \\

    \bottomrule
  \end{tabular}
  \caption{\textbf{PathVQA closed yes/no results} ($n{=}3{,}362$).
  VEA Acc evaluates the visual-evidence-absent control against the original label. $\Delta m_{\mathrm{VEA}}$ and retained margin measure confidence-level evidence dependence. ECE and Brier are computed on original images. Bold and underline mark the best and second-best trained rows; Base is an unadapted reference and is not ranked.}
  \label{tab:pathvqa}
\end{table*}

\begin{figure}[!t]
  \centering
  \includegraphics[width=0.92\linewidth]{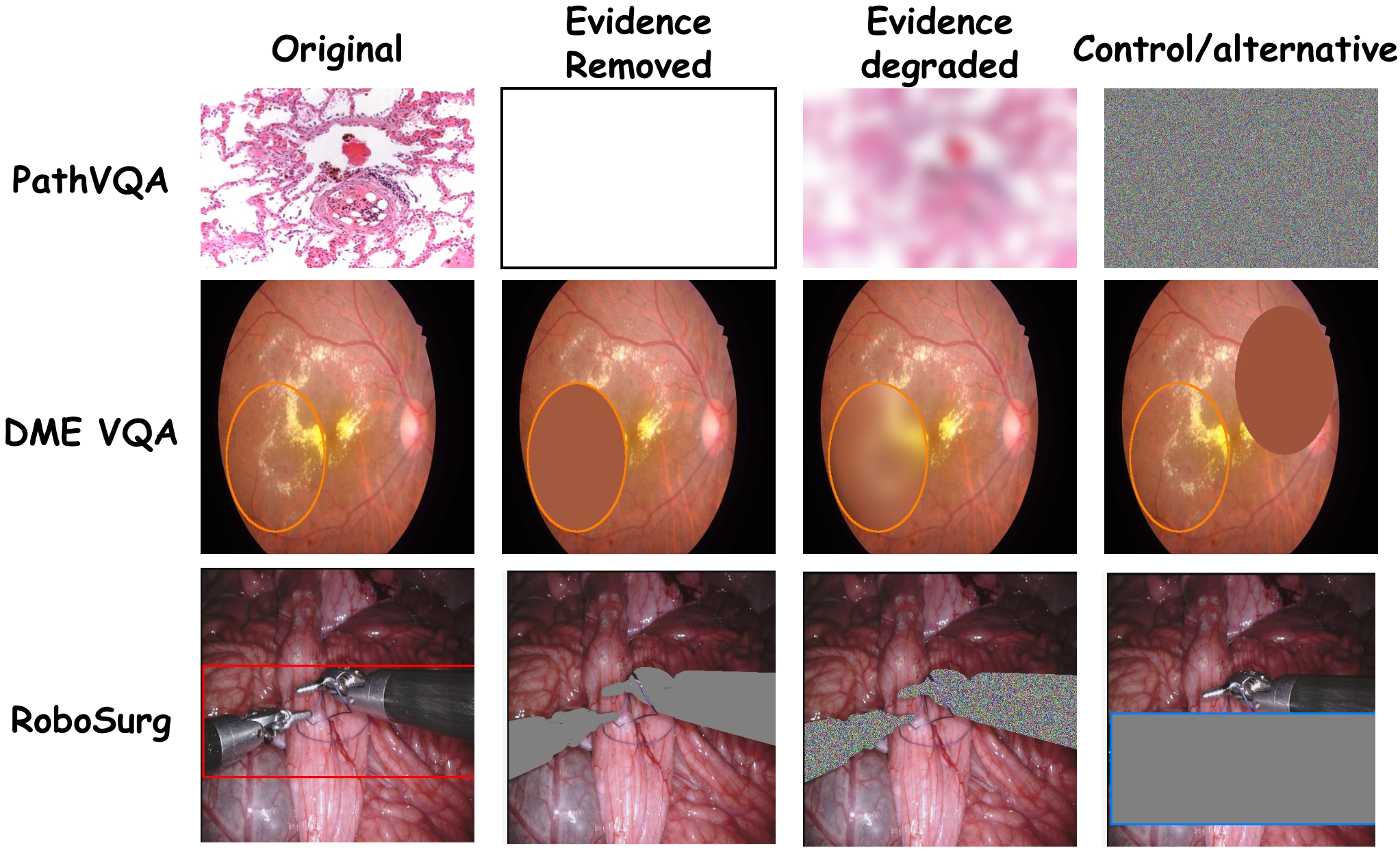}
  \caption{\textbf{Representative evidence-control construction.}
  The question and answer candidates stay fixed while visual evidence is removed
  or degraded at the relevant level: global for PathVQA, lesion-local for DME,
  and instrument-local for RoboSurg-VQA.}
  \label{fig:qual_controls}
\end{figure}

\paragraph{Evidence-control view.}
\Cref{fig:qual_controls} shows the three paired contrasts: global evidence for PathVQA, queried lesions for DME, and segmented instruments for RoboSurg-VQA. The tables test whether the response is stronger for relevant evidence than for matched controls.

\paragraph{Surgical instrument evidence.}
\label{sec:robosurg}
On RoboSurg-VQA, ECDD matches the highest original accuracy ($73.00\%$) and gives the largest instrument gap ($39.33$), positive margin drop ($1.811$), and instrument-minus-random response ($0.738$); no baseline matches this combination.

\paragraph{Global VEA support on PathVQA.}
\label{sec:pathvqa}
SFT and the five distillation baselines have VEA gaps of $4.85$--$6.28$ points. ECDD reaches $89.35\%$ original accuracy, a $20.26$-point gap, and $\Delta m_{\rm VEA}=3.863$, improving the strongest non-ECDD diagnostics by $14.0$ points and $3.11$ margin units. It also has the lowest original-image ECE/Brier ($1.38/15.19$).

\subsection{Behavior under Question Priors}
\label{sec:behavioral}

To define the subset independently of the image models, we train a Qwen-3B LoRA
question-only model for one epoch on the PathVQA training split using
fixed-candidate cross-entropy and no image input. Its frozen test accuracy is
$85.48\%$, and its 488 errors form the \emph{prior-conflicting} subset.

\begin{table}[!t]
  \centering
  \footnotesize
  \setlength{\tabcolsep}{4.2pt}
  \begin{tabular}{@{}lcc@{}}
    \toprule
    Method & Full Acc. $\uparrow$ & Conflict Acc. $\uparrow$ \\
    \midrule
    Question-only prior & 85.48 & -- (defines subset) \\
    SFT & 88.55 & 61.89 \\
    KD~\cite{hinton2015distilling} & 86.79 & 62.70 \\
    DBiLD~\cite{sun2026switchkd} & 84.00 & 59.84 \\
    MedAlign~\cite{chang2026enhancing} & 84.62 & 60.25 \\
    A3Tune~\cite{chang2025focus} & 85.22 & 54.30 \\
    Beta-KD~\cite{sun2026uncertainty} & 86.91 & 61.89 \\
    \textbf{ECDD (ours)} & \textbf{89.35} & \textbf{66.39} \\
    \bottomrule
  \end{tabular}
  \caption{\textbf{PathVQA prior-conflict evaluation.} The conflict subset is
  selected only by errors of the frozen question-only model. The ECDD row uses
the staged PathVQA recipe in \Cref{app:training_shared}.}
  \label{tab:path_prior_conflict}
\end{table}

ECDD exceeds SFT, KD, and DBiLD by $4.51$, $3.69$, and $6.56$ points on this
subset (paired $p{=}0.0021$, $0.0505$, and $0.0024$, respectively), and also
exceeds all three added baselines (all $p{<}0.024$).
This result describes the behavior of the complete staged ECDD model on
unmodified images when the learned question prior fails; it does not isolate
the contribution of the dependency loss. On morphology/histology questions, accuracy rises from
$94.88\%$ to $95.17\%$, while the
same-model original-to-question-only gap rises from $4.04$ to $24.03$ points.
Presence and Other accuracy gains are $1.22$ and $0.97$ points.

\subsection{Ablation under the Headline Objective}
\label{sec:target_ablation}

\begin{table}[!t]
  \centering
  \footnotesize
  \setlength{\tabcolsep}{2.6pt}
  \renewcommand{\arraystretch}{1.03}
  \resizebox{\columnwidth}{!}{%
  \begin{tabular}{@{}lcccc@{}}
    \toprule
    Variant & Orig. $\uparrow$ & ROI Gap $\uparrow$ & Rand. Gap $\to0$ & ROI--Rand. $\Delta m_+\uparrow$ \\
    \midrule
    Matched KD & \textbf{80.50} & 3.81 & \textbf{-0.11} & 0.786 \\
    Constant lower bound & \underline{80.39} & 7.52 & \underline{-0.54} & 0.971 \\
    Two-sided matching & 80.17 & \underline{15.47} & \underline{0.54} & \underline{1.755} \\
    Random-region endpoint & 79.96 & 2.40 & -0.65 & 0.747 \\
    \textbf{ECDD (ours)} & \underline{80.39} & \textbf{16.88} & 0.87 & \textbf{2.023} \\
    \bottomrule
  \end{tabular}%
  }
  \caption{\textbf{DME ablation under the headline objective.} All rows
  use the same joint CE+KD task objective, optimization rule, and training
  budget. Only the dependency target or endpoint changes.}
  \label{tab:dme_target_ablation}
\end{table}

The three controls remove teacher-conditioned strength, one-sidedness, and
question--region alignment. ECDD gives the largest specific response while
remaining within $0.11$ accuracy points of matched KD. Single-factor analysis
places its parameters in a moderate operating region (\Cref{app:sensitivity}).

\subsection{Evidence Specificity and Intervention Transfer}
\label{sec:diagnostics}

\begin{figure}[!ht]
  \centering
  \includegraphics[width=\linewidth]{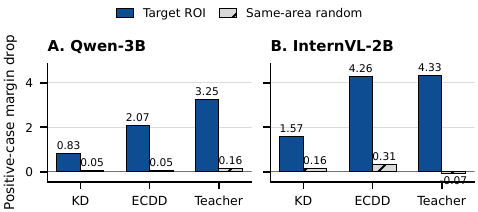}
  \caption{\textbf{Target-region evidence specificity.} On the DME lesion-presence
  VQA, bars show positive-case gold-margin drop under a target ROI mask or a
  same-area random mask. Across Qwen-3B and
  InternVL-2B, ECDD increases the target-ROI response over matched KD
  while keeping the random-mask response small. Teacher is an untuned diagnostic
  reference.}
  \label{fig:dme_region_dependency}
\end{figure}

\Cref{fig:dme_region_dependency} shows target specificity rather than generic
occlusion sensitivity. Across low/middle/high teacher-dependency tertiles,
ECDD's target-ROI margin drop rises $1.56/1.97/2.68$ (KD:
$0.50/0.77/1.24$), arguing against a uniform endpoint flip (\Cref{app:control_saturation}). We next replace the zero-filled training control with
Telea inpainting only at test time, while applying the same operator to a
shifted, same-area random region.

\begin{table}[!ht]
  \centering
  \footnotesize
  \setlength{\tabcolsep}{3.2pt}
  \resizebox{\columnwidth}{!}{%
  \begin{tabular}{@{}lrrrr@{}}
    \toprule
    Method & Orig. Acc. $\uparrow$ & Target $\Delta m_+\uparrow$ & Random $\Delta m_+\to0$ & Specific $\Delta m_+\uparrow$ \\
    \midrule
    SFT & 62.96 & 0.642 & -0.103 & 0.744 \\
    Matched KD & \textbf{80.50} & 0.746 & -0.032 & 0.778 \\
    \textbf{ECDD} & 80.39 & \textbf{1.167} & \textbf{-0.033} & \textbf{1.200} \\
    \bottomrule
  \end{tabular}%
  }
  \caption{\textbf{Transfer to Telea inpainting on DME.}
  Orig. Acc. repeats the reported original-image evaluation in
  \Cref{tab:dme_roi}; the remaining columns evaluate the frozen
  checkpoints with Telea inpainting. ECDD saw zero-filled ROI controls during
  training, whereas SFT and KD did not use evidence-control supervision. The
  ECDD gain over KD in specific response is $+0.423$ with an image-cluster bootstrap
  95\% CI of $[0.378,0.467]$.}
  \label{tab:dme_inpainting_main}
\end{table}

As shown in \Cref{tab:dme_inpainting_main}, ECDD retains KD-level original
accuracy while increasing the target-specific response and keeping the
matched random response near zero. On DME, the response therefore transfers
beyond the zero-filled control used during training.

\subsection{Limitations}
Our controls measure evidence dependency under specified interventions. They
do not establish clinical robustness or causal grounding. DME's blur and Telea
results support local intervention transfer, but do not establish equivalent
transfer for global PathVQA controls. Teacher-correctness filtering reduces
unreliable targets without excluding all teacher biases or intervention
artifacts. On RoboSurg-VQA, the dependency gain is smaller and random masking
also substantially affects accuracy; ECDD does not lead every localization
metric. The staged PathVQA recipe includes an additional training epoch and a
reference KL, so its task and calibration gains cannot be attributed to the
dependency loss alone. Finally, the present comparisons do not include ordinary
KD applied to both original and evidence-removed inputs. The current ablations
therefore support the design choices within ECDD, while leaving this
matched-input comparison open.

\section{Conclusion}
\label{sec:conclusion}
We introduced ECDD to transfer how answer support changes under a matched evidence intervention alongside the teacher's original-image preference. Original-image accuracy and evidence dependency describe different aspects of model behavior. Accuracy records whether the selected answer is correct, while the paired controls show how its support responds when relevant visual evidence is removed. Across lesion, instrument, and global controls, ECDD improves evidence-specific responses while retaining competitive accuracy. In the localized tasks, target regions produce a larger response than matched random regions, concentrating the effect on question-relevant evidence. ECDD changes training, not deployment: the student retains the same architecture, parameter count, and input as KD. Distillation for compact medical VLMs should preserve when an answer depends on visual evidence, not only which answer is selected.

\FloatBarrier
{\small
\bibliographystyle{ieeenat_fullname}
\bibliography{main}

@article{lau2018dataset,
  title={A dataset of clinically generated visual questions and answers about radiology images},
  author={Lau, Jason J and Gayen, Soumya and Ben Abacha, Asma and Demner-Fushman, Dina},
  journal={Scientific data},
  volume={5},
  number={1},
  pages={180251},
  year={2018},
  publisher={Nature Publishing Group}
}

@inproceedings{ben2019vqa,
  title={Vqa-med: Overview of the medical visual question answering task at imageclef 2019},
  author={Ben Abacha, Asma and Hasan, Sadid A and Datla, Vivek V and Demner-Fushman, Dina and M{\"u}ller, Henning},
  booktitle={Proceedings of CLEF (Conference and Labs of the Evaluation Forum) 2019 Working Notes},
  year={2019},
  organization={9-12 September 2019}
}

@inproceedings{huang2026bayesvqa,
  title={BayesVQA: Energy-Guided Bayesian Debiasing for Language-Bias-Robust Visual Question Answering},
  author={Huang, Zhiqi and Zhu, Huanjia and Deng, Xiangwen and Qinghao, Zhong and Chen, Bingzhi},
  booktitle={Proceedings of the AAAI Conference on Artificial Intelligence},
  volume={40},
  number={7},
  pages={5221--5229},
  year={2026}
}

@article{zhu2026language,
  title={Language-Bias-Resilient Visual Question Answering via Adaptive Multi-Margin Collaborative Debiasing},
  author={Zhu, Huanjia and Zheng, Shuyuan and Liu, Yishu and Cai, Sudong and Chen, Bingzhi},
  journal={Advances in Neural Information Processing Systems},
  volume={38},
  pages={125744--125763},
  year={2026}
}

@inproceedings{goyal2017making,
  title={Making the v in vqa matter: Elevating the role of image understanding in visual question answering},
  author={Goyal, Yash and Khot, Tejas and Summers-Stay, Douglas and Batra, Dhruv and Parikh, Devi},
  booktitle={Proceedings of the IEEE conference on computer vision and pattern recognition},
  pages={6904--6913},
  year={2017}
}

@inproceedings{agrawal2018don,
  title={Don't just assume; look and answer: Overcoming priors for visual question answering},
  author={Agrawal, Aishwarya and Batra, Dhruv and Parikh, Devi and Kembhavi, Aniruddha},
  booktitle={Proceedings of the IEEE conference on computer vision and pattern recognition},
  pages={4971--4980},
  year={2018}
}

@article{ramakrishnan2018overcoming,
  title={Overcoming language priors in visual question answering with adversarial regularization},
  author={Ramakrishnan, Sainandan and Agrawal, Aishwarya and Lee, Stefan},
  journal={Advances in neural information processing systems},
  volume={31},
  year={2018}
}

@inproceedings{cai2025llava,
  title={Llava-kd: A framework of distilling multimodal large language models},
  author={Cai, Yuxuan and Zhang, Jiangning and He, Haoyang and He, Xinwei and Tong, Ao and Gan, Zhenye and Wang, Chengjie and Xue, Zhucun and Liu, Yong and Bai, Xiang},
  booktitle={Proceedings of the IEEE/CVF International Conference on Computer Vision},
  pages={239--249},
  year={2025}
}

@article{zhang2023pmc,
  title={Pmc-vqa: Visual instruction tuning for medical visual question answering},
  author={Zhang, Xiaoman and Wu, Chaoyi and Zhao, Ziheng and Lin, Weixiong and Zhang, Ya and Wang, Yanfeng and Xie, Weidi},
  journal={arXiv preprint arXiv:2305.10415},
  year={2023}
}

@inproceedings{hu2024omnimedvqa,
  title={Omnimedvqa: A new large-scale comprehensive evaluation benchmark for medical lvlm},
  author={Hu, Yutao and Li, Tianbin and Lu, Quanfeng and Shao, Wenqi and He, Junjun and Qiao, Yu and Luo, Ping},
  booktitle={Proceedings of the IEEE/CVF Conference on Computer Vision and Pattern Recognition},
  pages={22170--22183},
  year={2024}
}

@article{hinton2015distilling,
  title={Distilling the knowledge in a neural network},
  author={Hinton, Geoffrey and Vinyals, Oriol and Dean, Jeff},
  journal={arXiv preprint arXiv:1503.02531},
  year={2015}
}

@inproceedings{shu2025llava,
  title={Llava-mod: Making llava tiny via moe-knowledge distillation},
  author={Shu, Fangxun and Liao, Yue and Zhang, Lei and Zhuo, Le and Xu, Chenning and Zhang, Guanghao and Shi, Haonan and Dai, Weilong and Yu, Zhelun and He, Wanggui and others},
  booktitle={International Conference on Learning Representations},
  volume={2025},
  pages={9386--9404},
  year={2025}
}

@article{xu2024llavadi,
  title={Llavadi: What matters for multimodal large language models distillation},
  author={Xu, Shilin and Li, Xiangtai and Yuan, Haobo and Qi, Lu and Tong, Yunhai and Yang, Ming-Hsuan},
  journal={arXiv preprint arXiv:2407.19409},
  year={2024}
}

@inproceedings{chang2026enhancing,
  title={Enhancing Medical Large Vision-Language Models via Alignment Distillation},
  author={Chang, Aofei and Wang, Ting and Ma, Fenglong},
  booktitle={Proceedings of the AAAI Conference on Artificial Intelligence},
  volume={40},
  number={24},
  pages={19952--19960},
  year={2026}
}

@article{chu2023mobilevlm,
  title   = {{MobileVLM}: A Fast, Strong and Open Vision Language Assistant for Mobile Devices},
  author  = {Chu, Xiangxiang and Qiao, Limeng and Lin, Xinyang and Xu, Shuang and Yang, Yang and Hu, Yiming and Wei, Fei and Zhang, Xinyu and Zhang, Bo and Wei, Xiaolin and Shen, Chunhua},
  journal = {arXiv preprint arXiv:2312.16886},
  year    = {2023}
}

@article{zhou2024tinyllava,
  title   = {{TinyLLaVA}: A Framework of Small-Scale Large Multimodal Models},
  author  = {Zhou, Baichuan and Hu, Ying and Weng, Xi and Jia, Junlong and Luo, Jie and Liu, Xien and Wu, Ji and Huang, Lei},
  journal = {arXiv preprint arXiv:2402.14289},
  year    = {2024}
}

@article{jin2024efficient,
  title   = {Efficient Multimodal Large Language Models: A Survey},
  author  = {Jin, Yizhang and Li, Jian and Liu, Yexin and Gu, Tianjun and Wu, Kai and Jiang, Zhengkai and He, Muyang and Zhao, Bo and Tan, Xin and Gan, Zhenye and Wang, Yabiao and Wang, Chengjie and Ma, Lizhuang},
  journal = {arXiv preprint arXiv:2405.10739},
  year    = {2024}
}

@inproceedings{vo2026vision,
  title     = {Vision Language Models are Biased},
  author    = {Vo, An and Nguyen, Khai-Nguyen and Taesiri, Mohammad Reza and Dang, Vy Tuong and Nguyen, Anh Totti and Kim, Daeyoung},
  booktitle = {The Fourteenth International Conference on Learning Representations (ICLR)},
  year      = {2026},
  url       = {https://arxiv.org/abs/2505.23941}
}

@article{zafar2026beyond,
  title   = {Beyond Accuracy: Evaluating Visual Grounding in Multimodal Medical Reasoning},
  author  = {Zafar, Anas and Murali, Leema Krishna and Vashist, Ashish},
  journal = {arXiv preprint arXiv:2603.03437},
  year    = {2026}
}

@article{he2020pathvqa,
  title={Pathvqa: 30000+ questions for medical visual question answering},
  author={He, Xuehai and Zhang, Yichen and Mou, Luntian and Xing, Eric and Xie, Pengtao},
  journal={arXiv preprint arXiv:2003.10286},
  year={2020}
}

@inproceedings{chang2025focus,
  title     = {Focus on What Matters: Enhancing Medical Vision-Language Models with Automatic Attention Alignment Tuning},
  author    = {Chang, Aofei and Huang, Le and Boyd, Alex James and Bhatia, Parminder and Kass-Hout, Taha and Xiao, Cao and Ma, Fenglong},
  booktitle = {Proceedings of the 63rd Annual Meeting of the Association for Computational Linguistics (ACL)},
  pages     = {9357--9372},
  year      = {2025}
}

@article{sun2026switchkd,
  title   = {Switch-{KD}: Visual-Switch Knowledge Distillation for Vision-Language Models},
  author  = {Sun, Haoyi and Wang, Xiaoxiao and Mao, Ning and Wang, Qian and Mu, Lifu and Zheng, Wen and Wei, Tao and Chen, Wei},
  journal = {arXiv preprint arXiv:2604.14629},
  year    = {2026}
}

@inproceedings{sun2026uncertainty,
  title     = {Uncertainty-Aware Knowledge Distillation for Multimodal Large Language Models},
  author    = {Sun, Jingchen and Han, Shaobo and Patel, Deep and Kohno, Wataru and Jin, Can and Chen, Changyou},
  booktitle = {Proceedings of the IEEE/CVF Conference on Computer Vision and Pattern Recognition (CVPR)},
  year      = {2026},
  note      = {arXiv:2603.21426}
}

@article{peng2025overcoming,
  title   = {Overcoming Language Prior Problem in Medical Visual Question Answering via Knowledge Distillation},
  author  = {Peng, Wei and others},
  journal = {arXiv preprint arXiv:2501.05690},
  year    = {2025}
}

@article{bai2025qwen25vl,
  title={Qwen2.5-VL technical report},
  author={Bai, Shuai and Chen, Keqin and Liu, Xuejing and Wang, Jialin and Ge, Wenbin and Song, Sibo and Dang, Kai and Wang, Peng and Wang, Shijie and Tang, Jun and Zhong, Humen and Zhu, Yuanzhi and Yang, Mingkun and Li, Zhaohai and Wan, Jianqiang and Wang, Pengfei and Ding, Wei and Fu, Zheren and Xu, Yiheng and Ye, Jiabo and Zhang, Xi and Xie, Tianbao and Cheng, Zesen and Zhang, Hang and Yang, Zhibo and Xu, Haiyang and Lin, Junyang},
  journal={arXiv preprint arXiv:2502.13923},
  year={2025}
}

@inproceedings{tascon2022consistency,
  title={Consistency-preserving visual question answering in medical imaging},
  author={Tascon-Morales, Sergio and M{\'a}rquez-Neila, Pablo and Sznitman, Raphael},
  booktitle={International Conference on Medical Image Computing and Computer-Assisted Intervention},
  pages={386--395},
  year={2022},
  organization={Springer}
}

@article{zhang2026robosurg,
  title={RoboSurg-VQA: A Multimodal Benchmark for Surgical Segmentation-Aware Visual Question Answering},
  author={Zhang, Chengyi and Ye, Zi and Wang, Ziyang},
  journal={arXiv preprint arXiv:2605.23068},
  year={2026}
}

@article{allan20192017,
  title={2017 robotic instrument segmentation challenge},
  author={Allan, Max and Shvets, Alex and Kurmann, Thomas and Zhang, Zichen and Duggal, Rahul and Su, Yun-Hsuan and Rieke, Nicola and Laina, Iro and Kalavakonda, Niveditha and Bodenstedt, Sebastian and others},
  journal={arXiv preprint arXiv:1902.06426},
  year={2019}
}

@article{wang2025internvl3,
  title={Internvl3. 5: Advancing open-source multimodal models in versatility, reasoning, and efficiency},
  author={Wang, Weiyun and Gao, Zhangwei and Gu, Lixin and Pu, Hengjun and Cui, Long and Wei, Xingguang and Liu, Zhaoyang and Jing, Linglin and Ye, Shenglong and Shao, Jie and others},
  journal={arXiv preprint arXiv:2508.18265},
  year={2025}
}

@inproceedings{cadene2019rubi,
  title={RUBi: Reducing Unimodal Biases for Visual Question Answering},
  author={Cadene, Remi and Dancette, Corentin and Ben-younes, Hedi and Cord, Matthieu and Parikh, Devi},
  booktitle={Advances in Neural Information Processing Systems},
  year={2019}
}

@inproceedings{chen2020counterfactual,
  title={Counterfactual Samples Synthesizing for Robust Visual Question Answering},
  author={Chen, Long and Yan, Xin and Xiao, Jun and Zhang, Hanwang and Pu, Shiliang and Zhuang, Yueting},
  booktitle={Proceedings of the IEEE/CVF Conference on Computer Vision and Pattern Recognition},
  year={2020}
}

@inproceedings{niu2021counterfactual,
  title={Counterfactual VQA: A Cause-Effect Look at Language Bias},
  author={Niu, Yulei and Tang, Kaihua and Zhang, Hanwang and Lu, Zhiwu and Hua, Xian-Sheng and Wen, Ji-Rong},
  booktitle={Proceedings of the IEEE/CVF Conference on Computer Vision and Pattern Recognition},
  year={2021}
}

@article{lan2026performance,
  title={From performance to practice: knowledge-distilled segmentator for on-premises clinical workflows},
  author={Lan, Qizhen and Choi, Aaron and Ma, Jun and Wang, Bo and Zhao, Zhongming and Jiang, Xiaoqian and Hsu, Yu-Chun},
  journal={npj Health Systems},
  volume={3},
  pages={58},
  year={2026},
  doi={10.1038/s44401-026-00108-w},
  url={https://www.nature.com/articles/s44401-026-00108-w}
}

@inproceedings{fan2026dcd,
  title={Detail Consistent Stage-Wise Distillation for Efficient {3D MRI} Segmentation},
  author={Fan, Mengchen and Geng, Baocheng and Xiao, Xi and Wang, Tianyang and Mei, Siyuan and Che, Pulin and Jiang, Xiaoqian and Lan, Qizhen},
  booktitle={Medical Image Computing and Computer Assisted Intervention -- MICCAI 2026},
  publisher={Springer Nature Switzerland},
  year={2026},
  url={https://papers.miccai.org/miccai-2026/0265-Paper5877.html}
}

@inproceedings{ding2026dprd,
  title={Displacement Preserving Relational Distillation for Robust Medical Segmentation},
  author={Ding, Zhicheng and Chu, Xinyu and Choi, Jung Im and Tian, Qing and Shi, Tianyu and Jiang, Xiaoqian and Zhu, Lijing and Lan, Qizhen},
  booktitle={Medical Image Computing and Computer Assisted Intervention -- MICCAI 2026},
  publisher={Springer Nature Switzerland},
  year={2026},
  url={https://papers.miccai.org/miccai-2026/0293-Paper6226.html}
}

@article{lan2025clockdistill,
  title={{CLoCKDistill}: Consistent Location-and-Context-aware Knowledge Distillation for {DETRs}},
  author={Lan, Qizhen and Tian, Qing},
  journal={arXiv preprint arXiv:2502.10683},
  year={2025},
  url={https://arxiv.org/abs/2502.10683}
}

@inproceedings{lan2025acamkd,
  title={{ACAM-KD}: Adaptive and Cooperative Attention Masking for Knowledge Distillation},
  author={Lan, Qizhen and Tian, Qing},
  booktitle={Proceedings of the IEEE/CVF International Conference on Computer Vision (ICCV)},
  pages={3957--3966},
  year={2025},
  doi={10.1109/ICCV51701.2025.00377},
  url={https://openaccess.thecvf.com/content/ICCV2025/html/Lan_ACAM-KD_Adaptive_and_Cooperative_Attention_Masking_for_Knowledge_Distillation_ICCV_2025_paper.html}
}

@article{feng2026trimrg,
  title={{TriMRG}: Toward Hallucination-Resistant Medical Image Report Generation via Triple Graph Semantic Reasoning and Rethinking},
  author={Feng, Zehui and Zhu, Dian and Zhao, Jianan and Zhao, Peiquan and Ren, Jianing and Han, Ting},
  journal={IEEE Transactions on Circuits and Systems for Video Technology},
  year={2026},
  note={Early Access},
  doi={10.1109/TCSVT.2026.3712885},
  url={https://ieeexplore.ieee.org/document/11606415/}
}

@article{feng2026tfvsf,
  title={{TF-VSF}: A Novel Training-Free Visual--Semantic Fusion Rare Medical Morning Glory Syndrome Diseases Severity Assessment Method},
  author={Feng, Zehui and Shen, Chenghao and Wang, Pei and Zhao, Peiquan and Ren, Jianing and Han, Ting},
  journal={Journal of Imaging Informatics in Medicine},
  year={2026},
  doi={10.1007/s10278-026-01884-5},
  url={https://link.springer.com/article/10.1007/s10278-026-01884-5}
}
}
\clearpage
\appendix
\setcounter{table}{0}
\setcounter{figure}{0}
\renewcommand{\thetable}{A\arabic{table}}
\renewcommand{\thefigure}{A\arabic{figure}}
\raggedbottom
\section{Dataset Details}
\label{app:dataset_details}

\subsection{PathVQA: Global Visual Evidence}
\label{app:data_pathvqa}

We use the official PathVQA split~\citep{he2020pathvqa}. To keep the task
aligned with the candidate-scoring interface used by every method, we retain
questions whose ground-truth answer is exactly \emph{yes} or \emph{no} and use
the fixed candidates ``Yes'' and ``No''. This filtering yields 9,751 training
and 3,362 test examples. Open-answer questions are not included in this study.
The visual-evidence-absent (VEA) endpoint replaces the image with a same-size
blank control; the question, candidates, and gold answer remain unchanged.

\subsection{DME VQA: Region-Level Lesion Evidence}
\label{app:data_dme}

The Diabetic Macular Edema (DME) VQA dataset~\citep{tascon2022consistency}
provides fundus images, region-referential questions, and masks for the regions
named in those questions. Its public split contains 433/112/134
train/validation/test images and 9,779/2,380/1,311 QA pairs. We study the
closed-form questions that ask whether hard exudates are present in a specified
region, giving 8,498 training and 918 test examples after candidate conversion.

Each test item is evaluated with the original image, a question-specified ROI
mask, and a same-area random mask. The random mask provides the specificity
control: similar responses to the target and random masks would indicate generic
occlusion sensitivity, whereas a larger target-ROI response is consistent with
dependence on the region named by the question.

\subsection{RoboSurg-VQA: Scene-Local Instrument Evidence}
\label{app:data_robosurg}

We use RoboSurg-VQA~\citep{zhang2026robosurg} with the public EndoVis2017
instrument-segmentation frames~\citep{allan20192017}. The experiment focuses on
Q8, ``Is the center of the frame occluded by instruments?'', which has a closed
yes/no answer space and an explicit instrument-evidence interpretation. Site1
videos 1--8 provide 1,800 training examples; video10 provides 300 held-out test
examples. Video9 is omitted because the public preprocessed EndoVis2017 archive
is incomplete for that split. The target control masks instrument pixels using
the segmentation labels. Same-area random masking and instrument-noise controls
separate instrument-specific response from generic occlusion and low-level mask
artifacts.

\subsection{Annotation and Intervention Requirements}
The localized experiments reuse the released DME region definitions and
EndoVis2017 segmentation labels to construct masks programmatically. PathVQA's
blank endpoint does not require localized evidence annotations. Localized ECDD
therefore assumes access to the relevant region or object annotations; these
experiments do not estimate the cost of collecting such annotations in a new
deployment setting.

\begin{table}[H]
\centering\footnotesize
\begin{tabular}{@{}lll@{}}
\toprule
Setting & Training control & Evaluation controls \\
\midrule
DME & Target ROI zero mask & Target/random mask, blur, Telea \\
RoboSurg & Instrument zero mask & Instrument/random mask \\
PathVQA & Global blank & Blank, degraded/global controls \\
\bottomrule
\end{tabular}
\caption{Training endpoints and evaluation controls. DME blur and Telea
inpainting are held out from training. Random masks are specificity controls.}
\label{tab:intervention_usage}
\end{table}

\section{Metric Definitions}
\label{app:metric_details}

\subsection{Candidate and Endpoint Scoring}
\label{app:metric_scoring}

All original and control endpoints use the same option-letter prompt and the
same fixed candidate set $C_i$. For model $M$ under image condition $z$, the
option-token logit for candidate $k$ at the first answer position is
$r_M(k\mid z,q)$. We normalize over the candidate set only,
$p_M(k\mid z,q)=\exp r_M(k\mid z,q)/\sum_{\ell\in C_i}\exp r_M(\ell\mid z,q)$.
All subsequent margins use $s_M(k\mid z,q)=\log p_M(k\mid z,q)$. The original
and control endpoints therefore differ only in their visual input.

\subsection{Accuracy-Based Diagnostics}
\label{app:metric_accuracy}

For an original example $(x_i,q_i,y_i)$ and a control image $\tilde{x}_i=A(x_i)$,
control accuracy retains the original label,
$\mathrm{Acc}_A=n^{-1}\sum_i\mathbf{1}[\hat{y}_i(\tilde{x}_i,q_i)=y_i]$.
It is a behavioral diagnostic: the control image is not assigned a new clinical
ground truth. The corresponding gap is
$100(\mathrm{Acc}_{\mathrm{orig}}-\mathrm{Acc}_A)$ percentage points. A larger
gap is meaningful only when original-image accuracy is preserved.

\subsection{Margin-Based Diagnostics}
\label{app:metric_margin}

For example $i$ and model $M$, the gold margin is
$m_M(z,q_i)=s_M(y_i\mid z,q_i)-\max_{y\in C_i\setminus\{y_i\}}s_M(y\mid z,q_i)$.
The corresponding single-example dependency is
$D_M(x_i,q_i;A)=m_M(x_i,q_i)-m_M(A(x_i),q_i)$. Its dataset mean is
\begin{equation}
  \begin{aligned}
  \Delta m_{M,A}
  &= \frac{1}{n}\sum_i D_M(x_i,q_i;A) \\
  &= \frac{1}{n}\sum_i
  \left[m_M(x_i,q_i)-m_M(A(x_i),q_i)\right].
  \end{aligned}
\end{equation}
We also report the fraction of the original margin retained at the control
endpoint,
\begin{equation}
  \mathrm{Ret.}\,m_{M,A}
  = \frac{\sum_i m_M(A(x_i),q_i)}{\sum_i m_M(x_i,q_i)} ,
\end{equation}
where a lower value indicates that less original answer support survives the
intervention.

\subsection{Localized and Multi-Control Response}
\label{app:metric_localized}

For DME ROI yes/no, Evidence Ret.\ and Random Ret.\ are $\mathrm{Acc}_A$ under
the question-specified ROI mask and the same-area random mask, respectively. For
DME Region4-MCQ, Evidence Ret.\ masks the correct highlighted region, while
Random Ret.\ masks a non-answer highlighted region of the same format. Evidence
Gap and Random Gap subtract these original-label retentions from original-image
accuracy. Specific $\Delta m$ is the target-evidence margin drop minus the
matched random- or distractor-control margin drop. For localized presence tasks,
we report the positive-case average
$\Delta m_{M,A}^{+}=|P|^{-1}\sum_{i\in P}D_M(x_i,q_i;A)$, where $P$
contains examples whose original answer indicates the presence of the queried
evidence. For lesion localization, we average over the full test set. The
contrast separates a response to the queried evidence from a response to
masking in general.

For PathVQA multi-control ablations, Avg Gap is the mean of the per-control
accuracy gaps over blank, gray, noise, blur, block-shuffle, and image-swap
controls. Avg $\Delta m$ is the corresponding mean of $\Delta m_{M,A}$, and Ret.
$m$ is the mean retained-margin ratio across those controls. These averages are
reported only to test whether the observed behavior is specific to a blank
endpoint.

\subsection{Calibration and Paired Uncertainty}
\label{app:metric_statistics}

ECE and Brier score are computed from original-image candidate probabilities;
both are task-calibration diagnostics, and lower values indicate better
calibration. They are not used as evidence-removal metrics.

For the Qwen-3B DME lesion-presence block, paired bootstrap uses the same 918 test
examples for both methods ($B{=}5000$, seed 20260618). We compare ECDD with
DBiLD because it is the closest original-image-accuracy comparator in that
block. As \Cref{tab:dme_bootstrap_ci} shows, their accuracy difference is small,
whereas the paired intervals for both dependency diagnostics are positive.

\begin{table}[t]
  \centering
  \footnotesize
  \setlength{\tabcolsep}{3.2pt}
  \resizebox{\columnwidth}{!}{%
  \begin{tabular}{lccc}
    \toprule
    Metric & ECDD--DBiLD & 95\% CI & Test \\
    \midrule
    Orig Acc (pp) & $0.11$ & $[-0.87, 1.09]$ & McNemar $p{=}1.00$ \\
    ROI Gap (pp) & $12.20$ & $[9.59, 14.92]$ & bootstrap \\
    ROI--Rand.\ $\Delta m_+$ & $0.97$ & $[0.91, 1.03]$ & bootstrap \\
    \bottomrule
  \end{tabular}%
  }
  \caption{\textbf{Paired DME bootstrap deltas for the Qwen lesion-presence
  block.} Deltas are ECDD minus DBiLD on the same test examples. Accuracy is
  used as a preservation metric; the larger positive deltas are on the
  evidence-dependency diagnostics.}
  \label{tab:dme_bootstrap_ci}
\end{table}

\section{Training and Comparison Protocol}
\label{app:training_details}

\subsection{Training Settings}
\label{app:training_shared}

All adapted models use LoRA while the vision encoder remains frozen. Each training stage lasts one epoch, and teacher outputs are cached before
optimization. PathVQA uses two stages as specified below. ECDD adds one student endpoint during training but leaves the
inference architecture and input unchanged.

The Qwen-3B DME headline model jointly optimizes task supervision,
answer-distribution distillation, and ECDD in one stage with
$\lambda_{\mathrm{KD}}=1$, $\lambda_{\mathrm{ECDD}}=0.02$, $\rho=0.5$,
and $\tau=8$. The DME learning rate
is $10^{-5}$, with bfloat16 precision, gradient accumulation of 8, and a
teacher-reliability threshold of $\eta=0.5$. The remaining ECDD settings are
listed below.

\begin{table}[H]
  \centering
  \footnotesize
  \setlength{\tabcolsep}{3.2pt}
  \begin{tabular}{@{}lccc@{}}
    \toprule
    Setting & $\lambda_{\mathrm{ECDD}}$ & $\rho$ & $\tau$ \\
    \midrule
    DME, Qwen-3B       & 0.02 & 0.5 & 8 \\
    DME, InternVL-2B   & 0.05 & 0.7 & 8 \\
    RoboSurg-VQA, Qwen-3B & 0.03 & 0.5 & 4 \\
    PathVQA, Qwen-3B (continuation) & 0.05 & 1.0 & 8 \\
    \bottomrule
  \end{tabular}
  \caption*{\textbf{ECDD settings used for the reported comparison rows.}}
\end{table}

The reported PathVQA ECDD model starts from an SFT adapter trained for one
epoch, then trains for one further epoch with
$\mathcal L_{\rm CE}+0.05\mathcal L_{\rm ECDD}+0.1\mathcal L_{\rm anchor}$,
where $\mathcal L_{\rm anchor}=\mathrm{KL}(p_{S_0}(\cdot\mid x,q)\|p_S(\cdot\mid x,q))$
and $S_0$ is the frozen SFT model. Original-image teacher KD is disabled in this
continuation. Its original, VEA, prior-conflict, question-type, and multi-control
results all refer to this staged instantiation; the three-seed PathVQA ECDD
rows repeat the same staged recipe. The one-stage baselines do not receive
this extra epoch or reference term.

PathVQA uses a global visual-evidence-absent (VEA) endpoint. DME masks the
question-specified lesion region, and RoboSurg-VQA masks segmented instrument
pixels. For the localized tasks, a same-area random control tests whether the
response is specific to the evidence named by the question rather than to
occlusion alone.

\subsection{DME Training Configuration}
\label{app:training_dme_regimes}

For Qwen-3B DME, task learning, answer-level distillation, and dependency
transfer are optimized jointly with the objective
$\mathcal L_{\rm CE}+\mathcal L_{\rm KD}
+\lambda_{\rm ECDD}\mathcal L_{\rm ECDD}$ with these values. \Cref{tab:dme_target_ablation} holds this objective and training budget fixed while changing only the
dependency target or visual endpoint. All Qwen-3B DME lesion-presence
diagnostics evaluate the ECDD model reported in the main DME result. Matched KD
uses the same task recipe with $\lambda_{\mathrm{ECDD}}=0$.
For the Qwen-3B DME headline model, task and weighted dependency gradients are
computed separately. If their inner product is negative, the component of the
dependency gradient parallel to the task gradient is removed before the two
are added. This gradient-conflict projection preserves the task-gradient
component and is part of the reported optimization recipe.

For localized presence questions, $\mathcal{L}_{\mathrm{ECDD}}$ is active
when the queried evidence is present and the teacher provides a reliable positive
dependency target. Every example retains original-image task supervision.

The Qwen-32B teacher is evaluated without DME adaptation and reaches $69.39\%$
original-image accuracy. The Qwen-3B student is trained on 8,498 gold-labeled DME
questions and reaches $80.39\%$. During training, the teacher supplies the
original-image soft distribution and paired dependency target.

\subsection{Baseline Training and Computational Cost}
\label{app:training_cost}

The added MedAlign, A3Tune, and Beta-KD baselines use the same student and teacher
backbones, candidate-scoring interface, original images, and cached teacher
outputs. ECDD uses this setup together with the paired control endpoint and
dependency trace.

We profile the Qwen-3B DME training configuration on the same H100. After
64 warm-up examples, matched KD requires 286 ms/example and 9.91 GiB peak
allocated memory over 1,024 measured examples; ECDD requires 529 ms/example and
12.29 GiB. ECDD therefore requires $1.85\times$ the student-training time; this
comparison excludes the one-time generation of teacher traces. At inference, ECDD uses the same student
architecture, parameter count, and input as KD, so it introduces no additional
inference overhead.

\section{DME Mechanism Ablations under the Joint Objective}
\label{app:dme_ablation}

\Cref{tab:dme_target_ablation} fixes the task objective and varies only the
dependency target or visual endpoint. Constant lower bound, Two-sided matching,
and Random-region endpoint each remove one part of ECDD. All three produce lower
target-specific responses, supporting both the teacher-derived lower bound and
the intervention matched to the question.

\paragraph{Reference anchor control.}
The DME headline objective and its matched baselines omit the reference
anchor. To evaluate the anchor separately, we add
$0.1\mathcal L_{\rm anchor}$ to matched KD while keeping the remaining objective
and training settings fixed, where
$\mathcal L_{\rm anchor}=\mathrm{KL}(p_{S_0}(\cdot\mid x,q)\|p_S(\cdot\mid x,q))$
and $S_0$ is the frozen SFT model.

\begin{table}[H]
  \centering
  \footnotesize
  \setlength{\tabcolsep}{3.5pt}
  \resizebox{\columnwidth}{!}{%
  \begin{tabular}{@{}lrrrr@{}}
    \toprule
    Variant & Orig. Acc. & ROI Gap & Rand. Gap & ROI--Rand. $\Delta m_+$ \\
    \midrule
    Matched KD & 80.50 & 3.81 & -0.11 & 0.786 \\
    Matched KD + anchor & 80.28 & 4.14 & -0.76 & 0.781 \\
    \bottomrule
  \end{tabular}%
  }
  \caption{\textbf{Reference anchor control with the headline objective.}}
  \label{tab:dme_anchor}
\end{table}

The specific response changes from $0.786$ to $0.781$, while original-image
accuracy changes from $80.50\%$ to $80.28\%$.

\FloatBarrier
\section{Control Diagnostics and Operator Transfer}
\label{app:control_endpoints}

\subsection{Endpoint Prediction Distributions}
\label{app:endpoint_distributions}

\Cref{tab:dme_endpoint_distribution} reports prediction distributions for
the teacher, matched KD, and the ECDD model used in the main DME result. On blank
images, all three predict ``No'' for all 918 questions; the resulting $58.50\%$
accuracy equals the negative class prevalence ($537/918$). For ECDD, target
masking reduces the predicted-Yes rate to $5.45\%$, while the same-area random
mask leaves it at $48.04\%$.

\begin{table*}[t]
  \centering
  \scriptsize
  \setlength{\tabcolsep}{3.2pt}
  \resizebox{\textwidth}{!}{%
  \begin{tabular}{@{}lrrrrlr@{}}
    \toprule
    Endpoint & Teacher Pred. Yes & KD Pred. Yes & ECDD Pred. Yes & ECDD Acc. & ECDD confusion [YY, YN, NY, NN] & ECDD pos. margin median \\
    \midrule
    Original & 58.17 & 47.71 & 49.35 & 80.39 & [327, 54, 126, 411] & 1.000 \\
    Target ROI mask & 28.00 & 22.98 & 5.45 & 63.51 & [48, 333, 2, 535] & -1.250 \\
    Same-area random mask & 57.84 & 46.51 & 48.04 & 79.52 & [317, 64, 124, 413] & 1.000 \\
    Blank & 0.00 & 0.00 & 0.00 & 58.50 & [0, 381, 0, 537] & -0.625 \\
    ROI blur & 35.51 & 27.78 & 16.78 & 71.13 & [135, 246, 19, 518] & -0.250 \\
    \bottomrule
  \end{tabular}%
  }
  \caption{\textbf{DME endpoint prediction distributions for the headline ECDD model.}
  Percentages are computed over 918 test questions. YY and NN denote correct
  positive and negative predictions; YN and NY denote the two error directions.}
  \label{tab:dme_endpoint_distribution}
\end{table*}

\subsection{Dependency-Stratified Response}
\label{app:control_saturation}

We stratify the 381 positive cases into equal-size tertiles using the teacher's
target-ROI margin drop. Across the low-, middle-, and high-dependency groups,
ECDD's corresponding drops rise from $1.563$ to $1.971$ to $2.680$; matched KD
rises from $0.496$ to $0.765$ to $1.244$. At the group level, the student
response therefore follows the teacher's dependency strength.

\subsection{Held-Out ROI Blur}
\label{app:control_blur}

\begin{table}[t]
  \centering
  \footnotesize
  \setlength{\tabcolsep}{3.5pt}
  \begin{tabular}{@{}lrrrr@{}}
    \toprule
    Control & Teacher & SFT & Matched KD & ECDD \\
    \midrule
    Target ROI mask & 3.254 & 0.928 & 0.835 & 2.072 \\
    Same-area random mask & 0.161 & -0.064 & 0.049 & 0.049 \\
    Held-out ROI blur & 2.164 & 0.550 & 0.692 & 1.215 \\
    \bottomrule
  \end{tabular}
  \caption{\textbf{Positive-case gold-margin drop under DME controls.} Training
  uses the target ROI mask; ROI blur and Telea inpainting are held out for
  evaluation.}
  \label{tab:dme_blur_transfer}
\end{table}

At evaluation, ROI blur replaces target-region pixels with a Gaussian-blurred
copy of the same image (radius 12). ECDD retains a larger response than matched
KD ($1.215$ vs. $0.692$), while their same-area random responses coincide after
rounding ($0.049$ each). The target-specific separation therefore persists when
the region is blurred rather than zeroed.

\subsection{Held-Out Content-Aware Inpainting}
\label{app:control_inpainting}

We next evaluate the frozen checkpoints with OpenCV Telea content-aware
inpainting (radius 7), a held-out operator that replaces the training-time
zero-filled interior and fill discontinuity while retaining the same ROI
overlay. Its matched random control applies the same operator to a
deterministically shifted mask with the same shape and area, selected to
minimize overlap with the target ROI.

\begin{table}[t]
  \centering
  \footnotesize
  \setlength{\tabcolsep}{3.4pt}
  \resizebox{\columnwidth}{!}{%
  \begin{tabular}{@{}lrrrr@{}}
    \toprule
    Method & Orig. Acc. & Target $\Delta m_+$ & Random $\Delta m_+$ & Target--Random $\Delta m_+$ \\
    \midrule
    SFT & 62.96 & 0.642 & -0.103 & 0.744 \\
    Matched KD & \textbf{80.50} & 0.746 & -0.032 & 0.778 \\
    \textbf{ECDD} & 80.39 & \textbf{1.167} & \textbf{-0.033} & \textbf{1.200} \\
    \bottomrule
  \end{tabular}%
  }
  \caption{\textbf{Transfer to held-out content-aware ROI inpainting on DME.}
  Orig. Acc. repeats the reported original-image evaluation in \Cref{tab:dme_roi}; the margin responses use the frozen checkpoints under Telea. The
  random control uses the same inpainting operator and a shifted mask with the
  same geometry and area, selected to minimize target overlap.}
  \label{tab:dme_inpainting_transfer}
\end{table}

The evaluation contains 381 positive questions grouped into 23 image-name
clusters. An image-cluster bootstrap with 10,000 draws gives an ECDD--KD
difference of
$+0.423$ (95\% CI $[0.378,0.467]$) and an ECDD--SFT difference of $+0.456$
(95\% CI $[0.333,0.560]$). ECDD's random-region response remains near zero
($-0.033$). On DME, the target-region effect therefore persists under a
different removal operator, while the same operator at a random location
produces little response.

\section{Robustness across Training Seeds}
\label{app:multiseed}

We repeat SFT, KD, and ECDD with seeds 7, 17, and 27 while holding each training
recipe fixed. DME KD uses the same $\lambda_{\rm KD}=1.0$ task objective as ECDD
with $\lambda_{\mathrm{ECDD}}=0$. \Cref{tab:multiseed} reports mean $\pm$ sample
standard deviation. The dependency response is target-minus-random positive-case
margin drop for DME and RoboSurg-VQA, and original-to-VEA margin drop for PathVQA.
The metrics share the same interpretation, although their physical endpoints
differ. PathVQA additionally differs in training schedule: its ECDD row uses
the two-stage recipe, while SFT and KD use one epoch each.

\begin{table}[t]
  \centering
  \scriptsize
  \setlength{\tabcolsep}{2.2pt}
  \begin{tabular}{@{}llrrr@{}}
    \toprule
    Dataset & Method & \makecell{Orig.\\Acc.} & \makecell{Dependency\\response} & \makecell{ECDD--KD\\gain} \\
    \midrule
    DME & SFT & $63.91\pm1.12$ & $1.176\pm0.218$ & -- \\
        & KD & $80.46\pm0.82$ & $0.740\pm0.051$ & -- \\
        & ECDD & $80.46\pm0.91$ & $1.954\pm0.101$ & $1.213\pm0.054$ \\
    \midrule
    PathVQA & SFT & $88.65\pm0.07$ & $0.848\pm0.122$ & -- \\
            & KD & $86.87\pm0.21$ & $0.444\pm0.065$ & -- \\
            & ECDD & $89.81\pm0.40$ & $4.105\pm0.297$ & $3.661\pm0.241$ \\
    \midrule
    RoboSurg-VQA & SFT & $61.44\pm10.67$ & $0.481\pm0.026$ & -- \\
             & KD & $71.44\pm1.35$ & $0.509\pm0.082$ & -- \\
             & ECDD & $71.78\pm1.07$ & $0.634\pm0.109$ & $0.125\pm0.026$ \\
    \bottomrule
  \end{tabular}
  \caption{\textbf{Three-seed robustness.} Reported gains are mean $\pm$
  sample standard deviation of the seed-wise paired ECDD--KD differences. The
  dependency-response gains are $+1.213\pm0.054$, $+3.661\pm0.241$, and
  $+0.125\pm0.026$ on DME, PathVQA, and RoboSurg-VQA, respectively. The
  corresponding original-accuracy differences are $0.00\pm0.29$,
  $+2.93\pm0.51$, and $+0.33\pm0.67$ points.}
  \label{tab:multiseed}
\end{table}

\section{Single-Factor Sensitivity}
\label{app:sensitivity}

\begin{figure*}[t]
  \centering
  \includegraphics[width=0.92\textwidth]{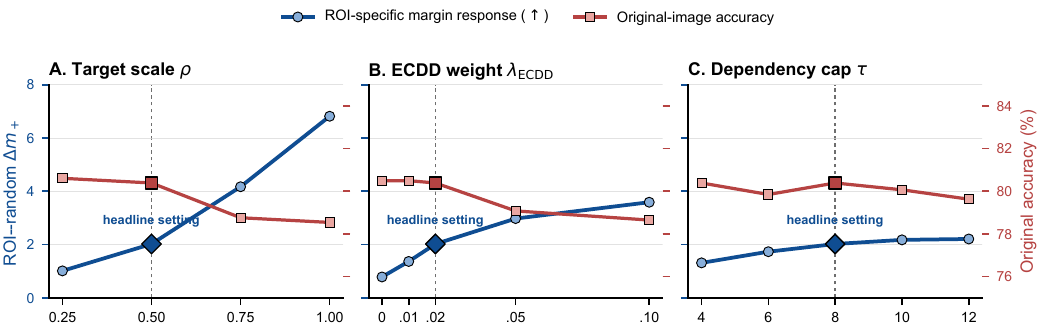}
  \caption{\textbf{Single-factor sensitivity for the DME training configuration.}
  Blue reports target-ROI minus random-region positive-case margin response;
  red reports original-image accuracy. Diamonds mark the headline setting.}
  \label{fig:dme_sensitivity}
\end{figure*}

The response changes smoothly around the selected setting. Increasing $\rho$
or $\lambda_{\mathrm{ECDD}}$ produces a gradual increase in ROI-specific
response, while original-image accuracy remains close to the selected setting
until the largest values. Raising
$\lambda_{\mathrm{ECDD}}$ from 0 to 0.01 increases the response from $0.786$ to
$1.368$ without changing $80.50\%$ accuracy; the headline value 0.02 reaches
$2.023$ with a $0.11$-point difference. The target-scale sweep follows the same
pattern, and the selected $\rho=0.5$ precedes the accuracy decline at the
largest setting. Across $\tau\in[4,12]$, accuracy varies by $0.76$ points, and
the response approaches a plateau at $\tau=8$--$12$. Together, the sweeps place
the headline values within a stable operating region.

\section{PathVQA Prior Diagnostics}
\label{app:path_prior}

\subsection{Question-Only Prior-Conflict Evaluation}
\label{app:path_prior_conflict}

We train a Qwen-3B LoRA question-only model for one epoch on the 9,751 PathVQA
training questions using fixed-candidate cross-entropy and no image input. Once
frozen, it reaches $85.48\%$ on the full test set. Its 488 errors define a
prior-conflict subset independently of every image model in the comparison. On
these original, unmodified images, ECDD
improves over SFT, KD, and DBiLD by $4.51$, $3.69$, and $6.56$ points. Paired
bootstrap 95\% intervals are $[1.84,7.38]$, $[0.20,7.38]$, and
$[2.46,10.66]$; exact McNemar $p$-values are $0.0021$, $0.0505$, and $0.0024$.
Using the same fixed subset and cached predictions, ECDD also improves over
MedAlign, A3Tune, and Beta-KD by $6.15$, $12.09$, and $4.51$ points; the paired
$p$-values are $0.0039$, $<10^{-8}$, and $0.0230$.

\subsection{Question-Type Analysis}
\label{app:path_question_types}

We use fixed, mutually exclusive text rules. Presence matches ``is/are
\emph{X} present?'' templates; Morphology/histology includes questions about
microscopy, stain, histology, nuclei, cells, or whether an image ``shows'' a
finding; all remaining questions are assigned to Other. These are diagnostic
buckets rather than official PathVQA labels.

\begin{table}[H]
  \centering
  \footnotesize
  \setlength{\tabcolsep}{2.8pt}
  \resizebox{\columnwidth}{!}{%
  \begin{tabular}{@{}lrrrrr@{}}
    \toprule
    Type & $n$ & SFT Orig. & ECDD Orig. & SFT Q-only gap & ECDD Q-only gap \\
    \midrule
    Presence & 1,561 & 85.78 & 87.00 & 8.71 & 8.90 \\
    Morphology/histology & 1,386 & 94.88 & 95.17 & 4.04 & 24.03 \\
    Other & 415 & 77.83 & 78.80 & 10.84 & 11.81 \\
    \bottomrule
  \end{tabular}%
  }
  \caption{\textbf{PathVQA question-type behavior.} The Q-only gap subtracts
  the accuracy of the same trained VLM with its image input omitted from its
  original-image accuracy. This endpoint differs from both the blank-image VEA
  evaluation and the separately trained question-only model used to define the
  prior-conflict subset.}
  \label{tab:path_question_type}
\end{table}

In \Cref{tab:path_question_type}, morphology/histology shows the clearest
change: the same-model Q-only gap rises from $4.04$ to $24.03$ points. Presence
and Other show smaller changes.

\subsection{Multiple Visual Controls}
\label{app:path_multicontrol}

\begin{figure}[H]
  \centering
  \includegraphics[width=\columnwidth]{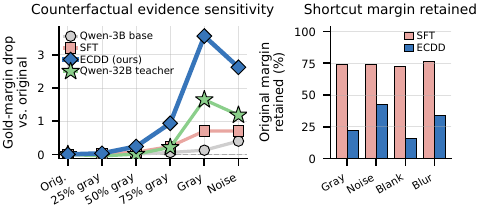}
  \caption{\textbf{PathVQA multi-control VEA diagnostic} ($n{=}3{,}362$).
  Lower retained margin means less original answer support survives the
  evidence-absent or degraded control. ECDD denotes the staged PathVQA model.}
  \label{fig:pathvqa_multicontrol_app}
\end{figure}

\Cref{fig:pathvqa_multicontrol_app} tests whether the global PathVQA result is
tied only to a blank endpoint. ECDD withdraws more gold-answer support than
Base, SFT, and the 32B teacher along the gray/noise degradation path. Across
gray, noise, blank, and blur controls, mean retained margin is $0.286$ for ECDD
and $0.745$ for SFT.

\FloatBarrier
\end{document}